\documentclass[10pt,twocolumn,letterpaper]{article}

\usepackage{cvpr}              

\usepackage[utf8]{inputenc} 
\usepackage[T1]{fontenc}    
\usepackage{url}            
\usepackage{booktabs}       
\usepackage{amsfonts}       
\usepackage{nicefrac}       
\usepackage{microtype}      
\usepackage{xcolor}         
\usepackage{multirow}

\usepackage{graphicx}
\usepackage{amsmath, amssymb}
\usepackage{algorithm}
\usepackage{algpseudocode}
\usepackage{multirow}
\usepackage[table]{xcolor}
\usepackage{tablefootnote}
\usepackage{subcaption}
\usepackage{wrapfig}
\usepackage{lipsum} 
\usepackage{booktabs} 
\usepackage{adjustbox}

\newcommand{\ModelOuput}{\mathbf{X}_{\theta}}

\definecolor{cvprblue}{rgb}{0.21,0.49,0.74}
\usepackage[pagebackref,breaklinks,colorlinks,allcolors=cvprblue]{hyperref}

\def\paperID{7786} 
\def\confName{CVPR}
\def\confYear{2026}

\title{Knot Forcing: Taming Autoregressive Video Diffusion Models for Real-time Infinite Interactive Portrait Animation}

\author{
    Steven Xiao\footnotemark[1] \quad Xindi Zhang\footnotemark[1] \quad Dechao Meng\footnotemark[1] \\ \quad Qi Wang \quad Peng Zhang \quad Bang Zhang \\
    Tongyi Lab, Alibaba Group 
}

\begin{document}
\maketitle
\renewcommand{\thefootnote}{\fnsymbol{footnote}}
\footnotetext[1]{Equal contribution.}

\begin{abstract}
Real-time portrait animation is essential for interactive applications such as virtual assistants and live avatars, requiring high visual fidelity, temporal coherence, ultra-low latency, and responsive control from dynamic inputs like reference images and driving signals. While diffusion-based models achieve strong quality, their non-causal nature hinders streaming deployment. Causal autoregressive video generation approaches enable efficient frame-by-frame generation but suffer from error accumulation, motion discontinuities at chunk boundaries, and degraded long-term consistency. In this work, we present a novel streaming framework named Knot Forcing for real-time portrait animation that addresses these challenges through three key designs: (1) a chunk-wise generation strategy with global identity preservation via cached KV states of the reference image and local temporal modeling using sliding window attention; (2) a temporal knot module that overlaps adjacent chunks and propagates spatio-temporal cues via image-to-video conditioning to smooth inter-chunk motion transitions; and (3) A ``running ahead" mechanism that dynamically updates the reference frame’s temporal coordinate during inference, keeping its semantic context ahead of the current rollout frame to support long-term coherence. Knot Forcing enables high-fidelity, temporally consistent, and interactive portrait animation over infinite sequences, achieving real-time performance with strong visual stability on consumer-grade GPUs. Project page: \url{https://humanaigc.github.io/knot_forcing_demo_page/}.


\end{abstract}    
\section{Introduction}
\label{sec:intro}

\begin{figure*}[t]
  \centering
   \includegraphics[width=0.92\linewidth]{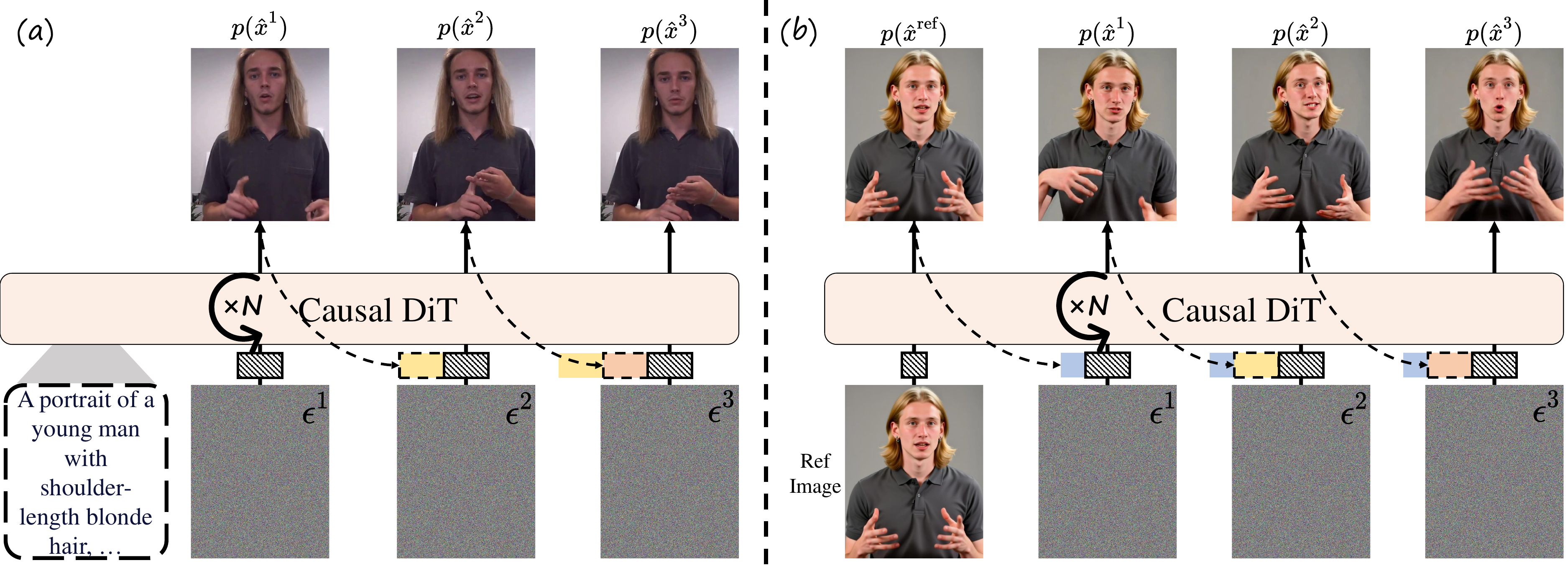}

   \caption{Streaming video generation setup. (a) T2V causal video diffusion. (b) Our approach for portrait animation: given a reference frame, we generate video autoregressively with a short sliding attention window, ensuring low latency, balanced computation, and stable identity preservation.}
   \vspace{-3mm}
   \label{fig:ref_swin}
\end{figure*}

Real-time portrait animation~\cite{guo2024liveportrait, xu2024vasa, meng2025mirrorme, zhen2025teller} has become a cornerstone of interactive digital experiences, powering applications such as virtual assistants, live avatars, and immersive communication systems. Recent advances in diffusion transformer (DiT)-based video generation~\cite{wan2025, liu2024sora, wan2025wan, kong2024hunyuanvideo, gao2025seedance, hacohen2024ltx} have provided a powerful engine for high-fidelity, photorealistic, and expressive digital human synthesis. Equipped with strong generative priors and sophisticated temporal modeling, these models support flexible conditioning on diverse user inputs—such as reference images, human poses, facial expressions, audio, or text—enabling highly controllable and personalized portrait animation~\cite{gao2025wan, cheng2025wan, lin2025omnihuman, jiang2025omnihuman, meng2025mirrorme}. However, these models typically operate on long token sequences and rely on iterative denoising, leading to high computational cost and slow generation speed. As a result, they struggle to meet the real-time demands of interactive applications, where low latency and immediate response are critical. This limits their practicality for on-the-fly portrait animation, despite their high visual quality.

Recently, causal autoregressive (AR) video generation model has emerged as a promising alternative for low-latency inference. By leveraging score distillation~\cite{yin2024improved, yin2025slow}, these works~\cite{yin2025slow, huang2025self, cui2025self, yang2025longlive, liu2025rolling, low2025talkingmachines} effectively transfer the rich contextual understanding of bidirectional teacher models into causal AR generators. Furthermore, by employing key-value (KV) caching and reducing the number of denoising steps, these models generate video frames in a streaming fashion (illustrated in Fig.~\ref{fig:ref_swin}(a)), significantly reducing real-time latency while preserving temporal coherence. However, these methods primarily focus on open-ended tasks such as text-to-video generation, and still suffer from noticeable visual degradation, flickering artifacts, and temporal inconsistency. 

In this work, we present $\text{Knot Forcing}$, a novel streaming framework tailored for real-time portrait animation, designed to balance quality, efficiency, and interactivity. Our approach integrates three key innovations: (1) We propose a chunk-wise causal generation framework that autoregressively produces video in manageable segments, balancing computational latency and temporal coherence, while supporting seamless integration of streaming control signals (e.g., poses or audio) for real-time controllability. To maintain identity consistency, we encode the user-provided reference image using the video DiT and cache its KV states as a global semantic anchor, which is fused with streaming video features throughout generation. A short sliding window attention is further applied to model local temporal dependencies between adjacent chunks (shown in Fig.~\ref{fig:ref_swin}(b)). (2) We observe that, unlike bidirectional models which provide each token with a consistent, full-sequence attention context, causal autoregressive video models suffer from periodic shifts in attention context between adjacent frames. This leads to visual drift and motion discontinuities at chunk boundaries, resulting in noticeable degradation in temporal stability and visual fidelity compared to the bidirectional teacher. To address this, we propose the Temporal Knot module, which introduces a temporal overlap between the tail frames of the previous chunk and the head frames of the current chunk. By leveraging image-to-video (I2V) conditioning, it propagates fine-grained spatio-temporal cues across chunk boundaries, effectively bridging semantic gaps and ensuring smooth motion transitions. This design enhances both local and global temporal consistency, enabling the causal model to better approximate bidirectional dynamics while preserving streaming efficiency and delivering high-fidelity portrait animation. (3) To address error accumulation in infinite portrait animation, we propose global context running ahead, which maintains a forward-looking goal state during synthesis. During training, the model learns from short clips with the final frame serving as a future-facing reference. At inference, we treat the ground-truth reference frame as a moving ``pseudo-final" frame, dynamically updating its temporal position according to the current generation step—by adjusting its rotary positional encoding (RoPE) and re-caching its KV states. This ensures the reference remains temporally ahead, providing a stable structural and identity prior that continuously guides streaming predictions toward the target trajectory. The approach effectively suppresses error propagation, preserves motion diversity, bridges the short-to-long gap between training and inference, and enhances visual fidelity in long-term generation.

We validate our method on diverse real-time, controllable portrait animation tasks under long-horizon setting. Experiments show that Knot Forcing outperforms existing approaches in visual stability, temporal coherence, and generation quality, with significantly reduced flickering, motion jitters, and identity drift. The framework maintains low latency and strong responsiveness to streaming controls, enabling high-fidelity, infinite-duration animation. Ablation studies confirm the contribution of each component, and overall, our method advances the state of the art in real-time, controllable portrait animation.

\section{Preliminaries}
\label{sec:pre}

\subsection{Autoregressive Video Diffusion} Current video diffusion transformers~\cite{liu2024sora, wan2025wan, kong2024hunyuanvideo, gao2025seedance, hacohen2024ltx} represent videos as extended token sequences, applying denoising across the full sequence. Although this yields videos with high coherence and quality, the high inference latency makes it difficult to adapt for real-time generation demands. Recently, the development of autoregressive (AR) video diffusion models~\cite{chen2025skyreels, teng2025magi, yang2025longlive, liu2025rolling, cui2025self, huang2025self, yin2025slow} offers a promising avenue for streaming generation. These models represent a hybrid approach, effectively combining the autoregressive chain-rule decomposition with the power of denoising diffusion models for video synthesis. Given a sequence of $N$ video frames $x^{1:N}=(x^1, x^2,...,x^N)$, AR models factorize the joint distribution into product of conditionals using the chain rule:

\begin{align}
    p(x^{1:N})={\textstyle \prod_{i=1}^{N}} p(x^i|x^{<i}),
    \label{eq:chain rule}
\end{align}

\noindent where each conditional distribution $p(x^i|x^{<i})$ is modeled via a diffusion process, wherein each frame $x^i$ is synthesized by progressively denoising Gaussian noise, conditioned on the preceding frames $x^{<i}$. Practically, many models~\cite{teng2025magi, yin2025slow} choose to model conditional distributions chunk by chunk, predicting $c$ future frames concurrently based on the current generated frames.


During inference, AR video diffusion models leverage KV caching to store the historical context $x^{<i}$ (shown in Fig.~\ref{fig:ref_swin}(a)). This allows for efficient reuse of past information when predicting subsequent frames, enabling streaming generation with reduced computational overhead.

\noindent\textbf{From Bidirectional to Autoregressive Video Diffusion.}  To build a robust Autoregressive (AR) video diffusion model, one promising approach is to distill knowledge from a pre-trained bidirectional teacher model $F_\phi(x_t,t)$ into a few-step causal model $G_\theta(x_t,t)$ via score matching~\cite{huang2025self,yin2025slow}. A Distribution Matching Distillation (DMD) loss~\cite{yin2024one, lu2025adversarial} is adopted to minimize the KL divergence between the target
distribution $p_\text{real}$ and the efficient generator output distribution $p_\text{fake}$. The gradient of DMD objective w.r.t. $\theta$ is:
\begin{equation}
    \begin{aligned}
    \nabla _\theta\mathcal{L}_{\text{DMD} }=\underset{z,t^{'},t,x_t}{\mathbb{E}} -[(s_\text{real}(x_t)-s_\text{fake}(x_t))\frac{\mathrm{d} G_\theta (z)}{\mathrm{d} \theta} ]
    \end{aligned}
\label{eq:DMD grad}
\end{equation}

\noindent where $z\sim \mathcal{N}(0, I)$, $t\sim \mathcal{U}(0,T)$, and $x_t=\Psi(G_\theta(z),t)$ is obtained by applying the forward diffusion process $\Psi$ to the output of $G_\theta$.


As formulated in Eq.~(\ref{eq:DMD grad}), $s_\text{real}(x_t)=\nabla_{x_t}\text{log}\ p_\text{real}(x_t)$ and $s_\text{fake}(x_t)=\nabla_{x_t}\text{log}\ p_\text{fake}(x_t)$ are score functions that point towards higher density for $p_\text{real}$ and $p_\text{fake}$, respectively.  Practically, $\nabla_{x_t}\text{log}\ p_\text{real}$ is estimated by the bidirectional teacher model $F_\phi(x_t,t)$, and $\nabla_{x_t}\text{log}\ p_\text{fake}$ is estimated by $f_\psi(x_t,t)$, which is initialized the same as $F_\phi(x_t,t)$, parameterized as a shadow model of $G_\theta$ within the continuous-time schedule, in order to provide distillation scores at multiple noise levels.

\noindent\textbf{Solving Train-inference Gap for AR Video Models.} During the inference phase, AR video diffusion models denoise each frame conditioned on their own generated past frames. Previous works such as Teacher Forcing and Diffusion Forcing~\cite{chen2024diffusion} train AR diffusion models by leveraging ground-truth prefixes (either clean or noised) to condition the denoising process of the next frame. However, such approaches introduce a significant train-inference gap. The mismatch between the training conditions (ground truth) and the inference conditions (model-generated history) results in significant video degradation and temporal inconsistency. To alleviate this, Self Forcing~\cite{huang2025self} directly sample training videos from the distributions estimated by the few-step $G_\theta$: $x_\theta^{1:N}=p_\theta(x^{1:N})={\textstyle \prod_{i=1}^{N}} p_\theta(x^i|x^{<i})$, where each frame $x_\theta^i$ is generated by iterative denoising, conditioned on KV-cached self-generated past clean frames and the current noisy frame. This approach ensures consistent exposure to self-generated prefix information during training, and drastically reduces the train-inference gap and minimizes artifacts in the generated videos.

\subsection{Controllable Portrait Animation}
\label{sec:controllable animation}
Controllable portrait animation~\cite{yang2025infinitetalk, wang2025fantasytalking, xue2025infinihuman, tian2024emo, hu2024animate, chen2025echomimic, lin2025omnihuman, jiang2025omnihuman, gao2025wan} is a challenging task that aims to generate realistic and consistent video sequences of a target person, driven by user-provided inputs. Achieving high visual fidelity in controllable portrait animation relies on two key aspects: maintaining the identity of the reference portrait image, and accurately reflecting the provided control signals (e.g., poses, audio, and motion dynamics). In this work, we adopt the Diffusion Transformer (DiT) architecture~\cite{wan2025wan} for portrait animation. Below we detail the fusion mechanisms for identity and driving signals injection respectively.

\noindent\textbf{ID Injection.} To inject the ID information from the user-provided reference image into the generated video, we encode the reference image using the video VAE. This static latent is then concatenated with the video latents along temporal dimension, enabling consistent identity fusion during generation~\cite{lin2025omnihuman, jiang2025omnihuman,kong2024hunyuanvideo}. We fine-tune a text-to-video diffusion model~\cite{gao2025wan} with masked inpainting to preserve identity, where an additional mask channel controls the visibility of the reference image. During training, we randomly expose reference image and past frames through the masks, teaching the model to reconstruct the target identity under varying conditions. At inference, only the reference image is visible.

\noindent\textbf{Driving Signals Injection.} Most user-provided control signals are abstract (e.g., audio, expression parameters, motion strength) and lack spatial alignment with video semantics. To ensure faithful adherence to such inputs, we introduce cross-attention layers~\cite{gao2025wan, meng2025mirrorme, lin2025omnihuman, jiang2025omnihuman} after selected DiT blocks, enabling frame-wise, content-aware fusion of driving signals with video features. The model learns the alignment between sequential video and corresponding driving signals from paired data, allowing it to condition on diverse control inputs and generate appropriately driven portrait animations. This design supports flexible, temporally coherent conditioning while maintaining identity and structural consistency.

\section{Method}

\begin{figure}[t]
  \centering
   \includegraphics[width=0.97\linewidth]{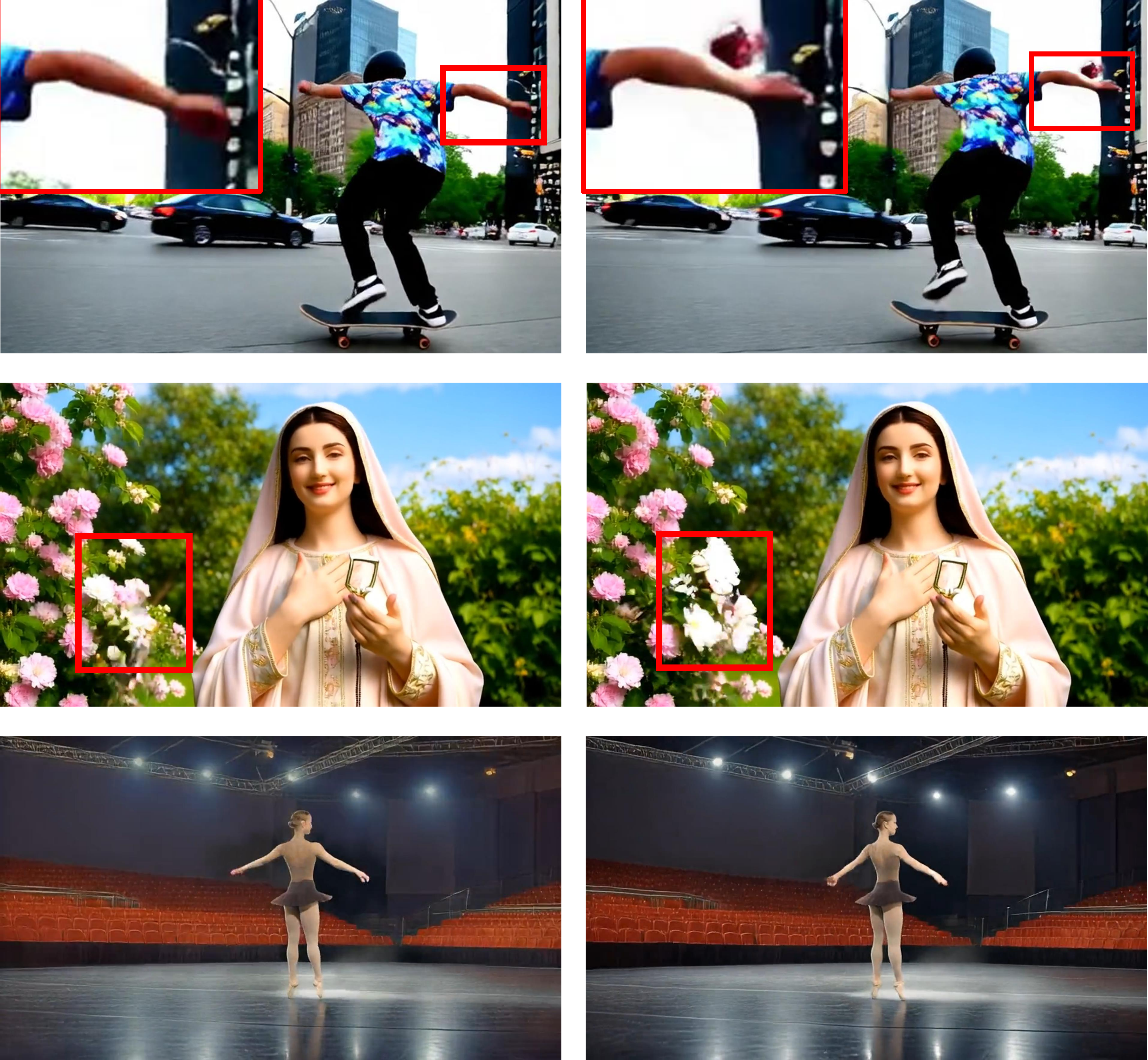}

   \caption{The video clips generated by Rolling Forcing, LongLive, and Self Forcing are presented from top to bottom, respectively. Significant temporal artifacts can be observed between adjacent frames, such as inconsistent object motion (first two rows) and abrupt changes in color tone (last row). Zoom in for details.}
   \label{fig:bad_causal}
   \vspace{-1mm}
\end{figure}

\subsection{Live Portrait Animation with Streaming Video Generation}
\label{sec:live portrait}

In practical applications, real-time portrait animation is highly desirable for scenarios such as virtual live streaming and visual dialogue, etc. In this work, we focus on devising a streaming generation method that achieves interactive portrait animation in real-time without compromising video fluency or quality.

\noindent\textbf{Stable AR Inference with Short Sliding Window and Global Context.} One straightforward approach is to apply Self Forcing for streaming portrait animation. Yet, the accumulation of memory tokens over time causes inference latency to grow, posing a challenge for real-time interactive generation that demands consistent and low latency. To achieve stable, low-latency inference, we introduce a short sliding window (Swin) of fixed length $L$. The sliding window limits attention to a local temporal context, ensuring constant per-chunk latency. However, restricting context may degrade long-range coherence and cause visual drift. To address this, we cache the KV pairs from the user-provided reference frame as a global anchor, preserving visual consistency. Under the noise schedule $\{t_0=0, t_1, ..., t_T=1000\}$ of the few-step $G_\theta$, generation of each chunk can be factorized into a multi-step denoising process, hence we rewrite the conditional distribution in Eq.~\ref{eq:chain rule} as follows:

\begin{equation}
    \begin{aligned}
        &p_\theta (x_{t_{j-1}}^{i:i+c}|x_{t_j}^{i:i+c},x_0^{i+c-L:i},x_0^{\text{ref} })\\=&\Psi(G_\theta(x_{t_j}^{i:i+c},t_{j},x_0^{i+c-L:i},x_0^{\text{ref} }),t_{j-1}),
\end{aligned}
\label{eq:ref_swin}
\end{equation}

\noindent where $t_{j-1},t_j\in\{t_0, t_1, ..., t_T\}$. At each denoising step for current chunk, generator $G_\theta$ predicts the clean frames conditioned on the noisy frames $x_{t_j}^{i:i+c}$, the local context within the sliding window, and the global context $x_0^{\text{ref}}$ from the reference image. Then the predicted output is converted to frames $x_{t_{j-1}}^{i:i+c}$ with reduced noise level via the forward process $\Psi$~\cite{liu2022flow}.

\begin{figure}[t]
  \centering
   \includegraphics[width=1.0\linewidth]{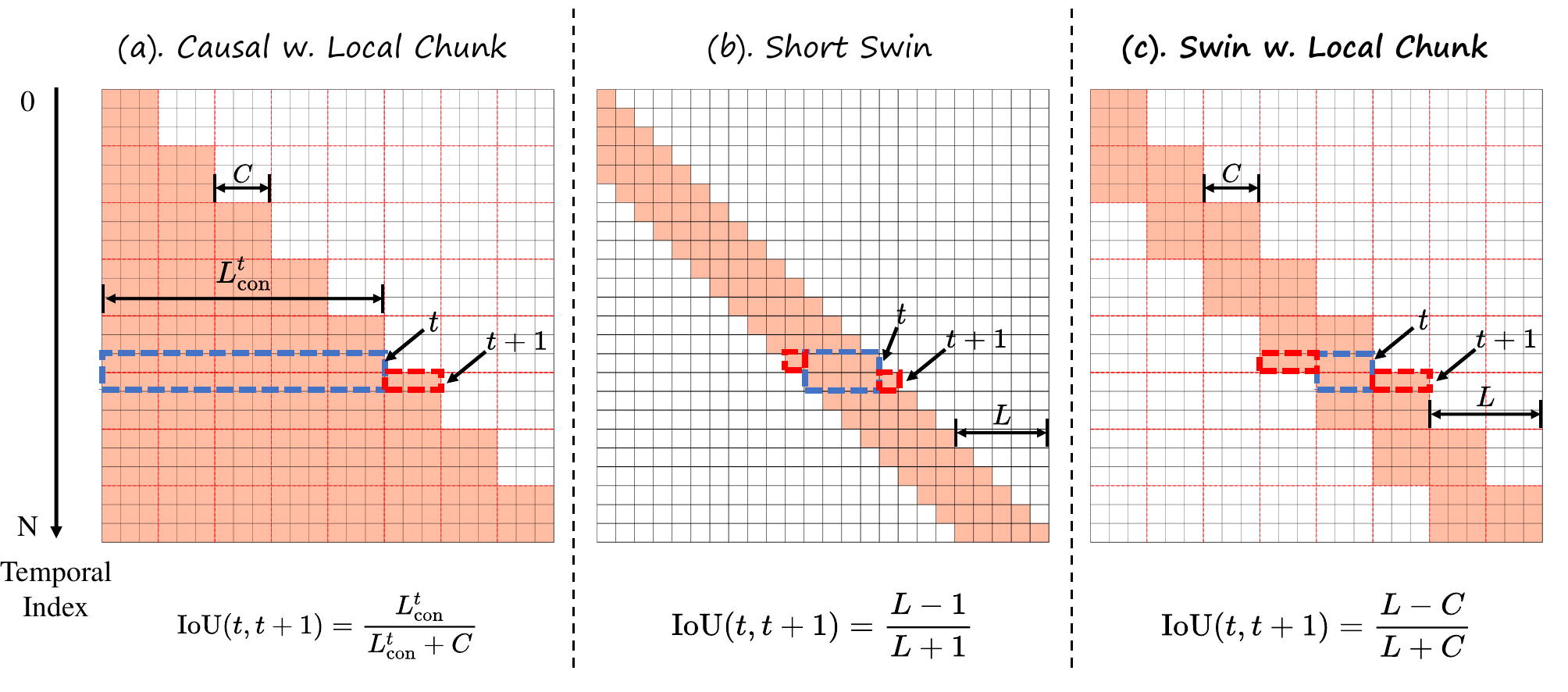}

   \caption{Attention masks of different causal designs. IoU of attention contexts between time steps $t$ and $t+1$ is computed to quantify the change in contextual coherence. (a): Causvid and Self Forcing. (b): LongLive. (c): Ours.}
   \label{fig:attn_context}
   \vspace{-3mm}
\end{figure}

\subsection{Forcing Inter-frame Coherence with Temporal Knot}
\label{sec: Knot Forcing}

\begin{figure*}[t]
  \centering
   \includegraphics[width=0.95\linewidth]{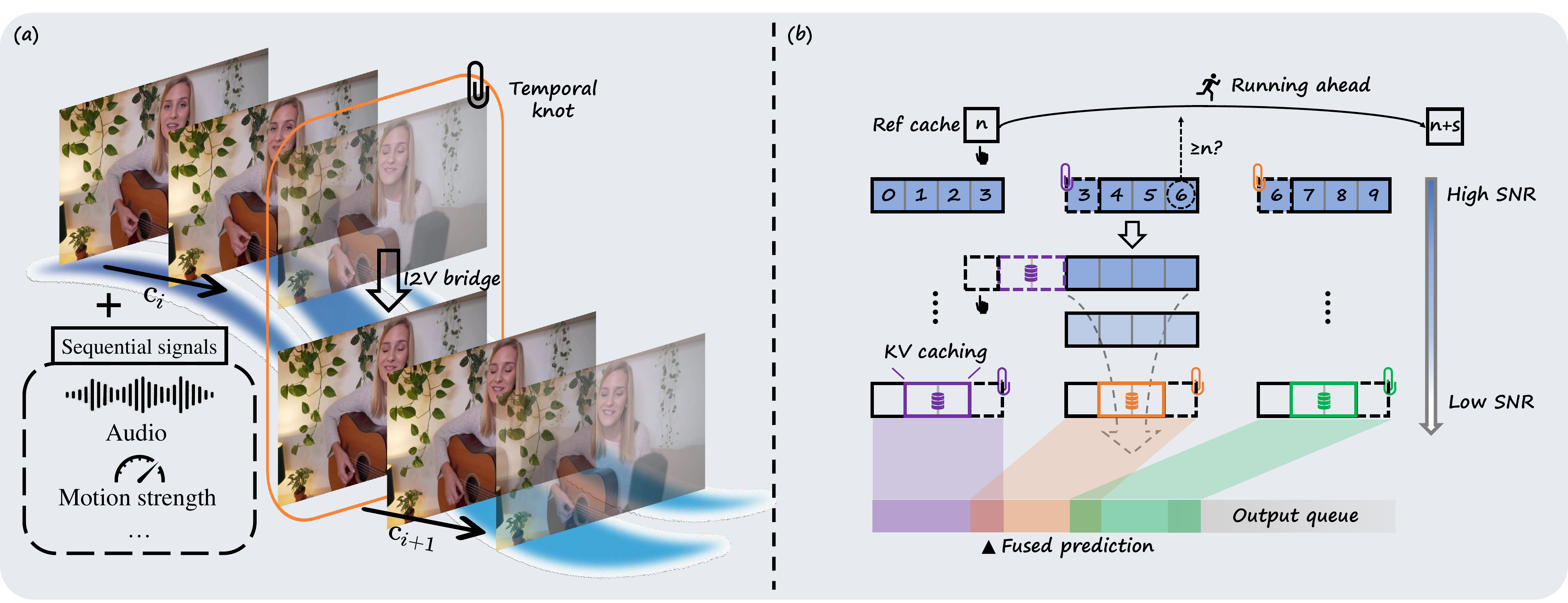}

   \caption{Key components of Knot Forcing. (a) illustrates the proposed temporal knot module. (b) illustrates the rollout inference pipeline with global context running ahead.}
   \label{fig:method}
    \vspace{-3mm}
\end{figure*}

 A critical challenge observed in the implementation of causal video diffusion models is the emergence of temporal drift between consecutively generated frames. This instability manifests as periodic oscillations in background color tones, sudden deformations in object shapes, and abrupt shifts in the periodicity of subject motion (shown in Fig.~\ref{fig:bad_causal}). These artifacts severely degrade the visual coherence and realism of synthesized videos. 
 
\noindent\textbf{Temporal Drift from Context Mismatch.} We identify the root cause of this issue as abrupt shifts in attention context between adjacent frame latents in causal generation. As shown in Fig.~\ref{fig:attn_context}, we visualize attention masks under various causal architectures and measure the similarity of attention contexts between neighboring frames. Our analysis reveals that all existing causal designs exhibit periodic shifts in attention context. This structural behavior fundamentally differs from the bidirectional teacher model $F_\phi(x_t,t)$, which leverages full temporal context to ensure stable and coherent latent transitions. In contrast, the autoregressive nature of the causal student $G_\theta$ limits each token’s receptive field to past frames only, making it difficult to approximate the teacher’s denoising distribution accurately.

These limitations induce feature misalignments that manifest as visible artifacts, such as flickering, shape warping, or motion jitters even between consecutive frames. Unlike discrete token prediction in language models~\cite{vaswani2017attention}, video diffusion operates in a continuous latent space~\cite{rombach2022high} requiring fine-grained spatial and temporal precision. The context mismatches amplify small errors into visible artifacts, and without future context, smooth frame-to-frame transitions become difficult to maintain, degrading temporal coherence.



\noindent\textbf{Temporal Knot as Semantic Bridge}. To address this, we propose \textit{Knot Forcing} to fix inter-frame coherence: at each step, the model simultaneously denoises the current chunk and the first $k$ frames of the subsequent chunk, explicitly aligning their semantic and motion context. This enables the model to incorporate future motion cues when predicting the current chunk, enhancing temporal coherence through locally aware, cross-chunk context fusion.

\begin{figure}[b]
  \centering
   \includegraphics[width=0.92\linewidth]{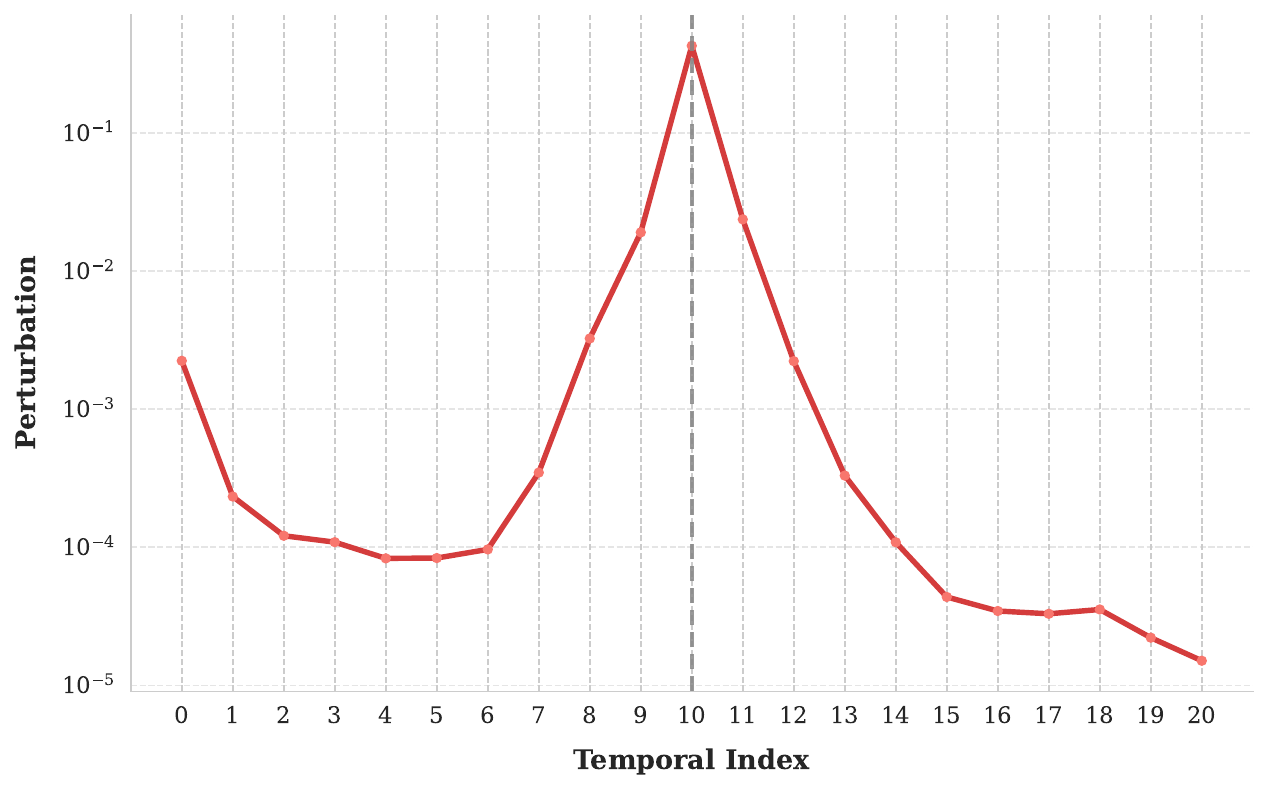}

   \caption{We assess inter-frame dependency by ablating each context frame and computing the L2 difference in attention outputs (relative to the original), normalized by the L2 norm of the unmodified output. The $10^\text{th}$ frame is used as the anchor, the resulting scores indicate each frame’s contribution to the current frame’s attention output.}
   \label{fig:mutual information}
\end{figure}

We refer to these shared boundary frames as temporal knots, which serve as anchor points that bind adjacent chunks together and stabilize transitions by maintaining consistent dynamics and semantics across generation steps. As shown in Fig.~\ref{fig:method}(a), inspired by Image-to-Video (I2V) video diffusion models~\cite{jiang2025vace}, we further propagate the early-predicted $k$ frames from the previous iteration to the next chunk via mask inpainting, ensuring identity-preserving and smooth frame transitions at chunk boundaries. Equipped with temporal knots, we reformulate the denoising process of $G_\theta$ to explicitly incorporate future context through shared boundary predictions. Specifically, we rewrite the denoising distribution in Eq.~\ref{eq:ref_swin} as:

\begin{equation}
    \begin{aligned}
        &p_\theta (x_{t_{j-1}}^{i:i+c};x_{t_{j-1}}^{i+c:i+c+k}|x_{t_j}^{i:i+c+k},x_0^{i+c-L:i},x_0^{\text{ref} })\\=&\Psi(G_\theta(x_{t_j}^{i:i+c+k},t_{j},\underline{x}^{i:i+k}_0, x_0^{i+c-L:i},x_0^{\text{ref} }),t_{j-1}),
\end{aligned}
\label{eq:knot forcing}
\end{equation}

\noindent where $\hat{x}^{i:i+k}_0$ represents the ``temporal knots'' generated by the model during the noise prediction of the prefix chunk latents. Eq.~\eqref{eq:knot forcing} enables a \textit{hinge-style} latent propagation, where local chunks maintain bidirectional context awareness, while the temporal knots bridge the semantic gap between adjacent chunks via mask inpainting. This restores the interrupted inter-chunk information flow at boundaries and mimics the full temporal modeling of the bidirectional teacher. In this way, Knot Forcing greatly improves the quality and temporal coherence of generated video flow.

Practically, each temporal knot introduces an additional $k$ frames of context for chunk denoising, which causes extra latency overhead. As shown in Fig.~\ref{fig:mutual information}, we quantitatively analyze the mutual information between video frames and their contextual neighbors. The results clearly show that the most informative context comes from adjacent frames in terms of attention modeling. To balance performance and latency, we set the hyperparameter $k=1$, ensuring effective context propagation while minimizing computational delay.

Essentially, the streaming latents at chunk boundaries are denoised twice, once as the suffix $\underline{x}^{i:i+k}_0$ of the preceding chunk and once as the prefix $x_0^{i:i+k}$ of the subsequent chunk. To improve consistency at chunk boundaries, we fuse the two predictions by taking their average as the final output for the knot regions (shown in Fig.~\ref{fig:method}(b)):

\vspace{-2mm} 
\begin{equation}
    \begin{aligned}  x_0^{i:i+k}\leftarrow \frac{x_0^{i:i+k} + \underline{x}^{i:i+k}_0}{2}
    \end{aligned}
\label{eq:fused prediction}
\end{equation}

\begin{figure}[t]
    \begin{minipage}[t]{0.45\textwidth}
      \begin{algorithm}[H]
        \caption{Knot Forcing Inference with Global Context Running Ahead}
        \small
        \begin{algorithmic}[1]
          \Require $\textbf{KV}_\text{pre}$ cache of prefix context
          \Require $\textbf{KV}_\text{ref}$ cache of global context
          \Require Denoise timesteps $\{t_0, \dots, t_T\}$
          \Require Number of generated frames $M$, chunk size $c$, local window length $L$, reference image $x_\text{ref}$, RoPE index $n$ for $x_\text{ref}$, running ahead interleave $s$
          \Require AR diffusion model $G_\theta$ (returns KV embeddings via $G_\theta^\text{KV}$)
          \State Initialize model output $\ModelOuput \gets []$
          \State Initialize KV cache $\textbf{KV}_\text{pre} \gets []$
          \State Initialize KV cache $\textbf{KV}_\text{ref} \gets G_\theta^\text{KV}(x_\text{ref}; 0)$
          \State Initialize Temporal Knot $\underline{x} \gets []$
          \While{$i <  M$}
            \If{$i+c+1 > n$}
              \State $n \gets n+s$ \Comment{Update RoPE Index for $x_\text{ref}$}
              \State $\textbf{KV}_\text{ref} \gets G_\theta^\text{KV}(x_\text{ref}; 0)$ \Comment{Running Ahead}
            \EndIf
            \State Initialize $x^{i:i+c+1}_{t_T} \sim \mathcal{N}(0, I)$
            \For{$j = T, \dots, 1$}
              \State Set $\hat{x}^{i:i+c+1}_{0} \gets G_\theta(x^{i:i+c+1}_{t_j}; \underline{x}, t_j, \textbf{KV}_\text{pre},\textbf{KV}_\text{ref})$
              \If{$j = 1$}
                \Comment{Update Temporal Knot}
                \If{$i>0$}
                    \State $\hat{x}_0^i \gets \frac{\underline{x}+\hat{x}^{i}_{0}}{2}$ \Comment{Fused Prediction}
                \EndIf
                \State $\underline{x} \gets \hat{x}^{i+c}_{0}$ 
                \State $\ModelOuput{\texttt{.append}}(\hat x^{i:i+c}_{0})$
                \State $\textbf{KV}_\text{pre} \gets G_\theta^\text{KV}(\hat{x}^{i+2c-L:i+c}_{0}; 0, \textbf{KV}_\text{pre},\textbf{KV}_\text{ref})$
              \Else
                \State Sample $\epsilon \sim \mathcal{N}(0, I)$
                \State Set $x^{i:i+c+1}_{t_{j-1}} \gets \Psi(\hat{x}^{i:i+c+1}_0, \epsilon, t_{j-1})$
              \EndIf
            \EndFor
            \State $i \gets i+c$
          \EndWhile
          \State \Return $\ModelOuput$
        \end{algorithmic}
        \label{alg:inference}
      \end{algorithm}
    \end{minipage}
    \vspace{-1em}
\end{figure}

\subsection{Mitigating Error Accumulation in Long-Term Streaming Generation}

Existing causal video generation methods suffer from severe error accumulation, as their attention receptive fields are relatively limited. During streaming generation, small prediction errors propagate and accumulate over time, leading to progressive degradation in visual details such as texture and structure. This ultimately results in a breakdown of temporal consistency in long-term sequences. Although existing methods~\cite{liu2025rolling,yang2025longlive} propose using the initially generated frames as attention sinks to preserve global semantic consistency, the generated content still gradually drifts away from the global context as the temporal index increases.

\noindent\textbf{Global Context Running Ahead}. To address this issue, inspired by previous work~\cite{jiang2025omnihuman}, we propose a global context running ahead strategy. During training, we consistently treat the last frame of each sampled video clip as the global context. During inference, we leverage the ground-truth reference image as the ``pseudo last frame" and ensure that its RoPE index is always placed beyond the current generation chunk, i.e., in the future relative to the generated frames. An overview of the inference pipeline is illustrated in Fig.~\ref{fig:method}(b).

As a result, the model learns to perceive the reference as a future-appearing anchor, which provides a consistent directional signal throughout the streaming generation process. This effectively guides the generation toward long-term coherence and mitigates error accumulation by continuously aligning predictions with a fixed semantic target. To facilitate a clearer understanding of our approach, we present the pseudo code for causal inference in Algorithm~\ref{alg:inference}.

\section{Experiments}
\label{exp}

\subsection{Implementation}

We implement Knot Forcing based on Wan2.1-T2V-1.3B~\cite{wan2025wan}. We first incorporate a mask inpainting module into the model and finetune on a dataset of 70k collected portrait videos, thereby obtaining a reference-based external signal-driven video generation model. Subsequently, following Self Forcing~\cite{huang2025self}, we initialize the causal model with the pretrained weights and distill the bidirectional model into a 4-step autoregressive video diffusion model. For hyperparameters, we set the chunk size $c$ to 3, the local window size $L$ to 6, and the temporal knot length $k$ to 1.

\subsection{Long-term Portrait Animation}
\begin{figure*}[t]
  \centering
   \includegraphics[width=0.92\linewidth]{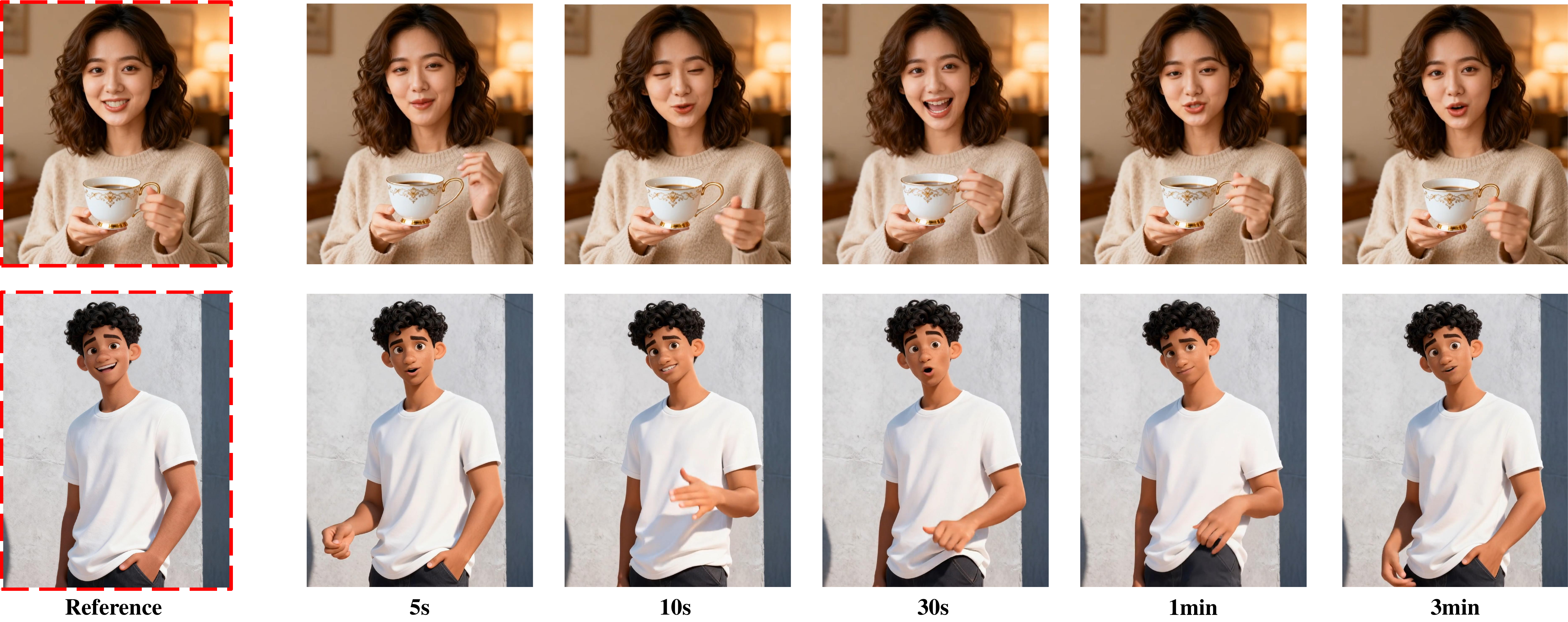}

   \caption{Infinite portrait animation results utilizing the proposed Knot Forcing. Given a reference frame, our approach produces videos exhibiting smooth motion, consistent identity, and high vividness across long-horizon without drift during stream-based generation. The generated timestamps are annotated beneath the corresponding video frames.}
   \label{fig:exp1}
   \vspace{-2mm}
\end{figure*}

As shown in Fig.~\ref{fig:exp1}, we evaluate the performance of our model on infinite portrait animation. As demonstrated, Knot Forcing effectively maintains both visual vividness and temporal stability during generation without introducing error accumulation. The proposed temporal knot successfully bridges the temporal gap between consecutive chunks in streaming generation. Furthermore, the proposed global context running ahead further provides the model with effective lookahead guidance, correcting temporal drift that may occur over extended sequences. 

\subsection{Comparisons}

\subsubsection{Compare with Autoregressive Portrait Animation}
\label{exp:comp_ar}

\begin{figure}[b]
  \centering
   \includegraphics[width=0.97\linewidth]{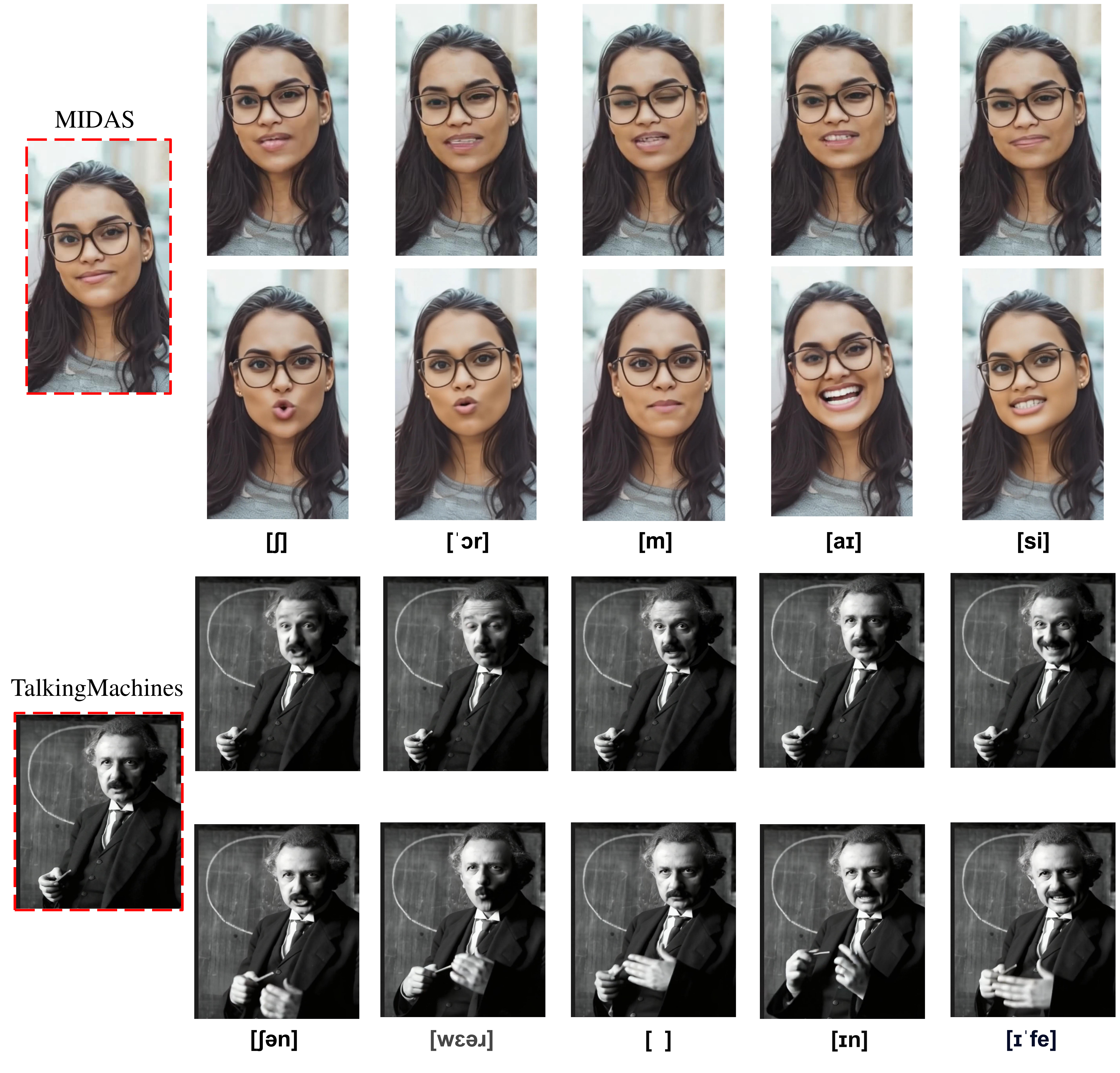}

   \caption{Comparisons with streaming portrait animation models. Phonemes corresponding to the video frames are annotated below.}
   \label{fig:exp_comp}
\end{figure}

We compare our proposed approach with two portrait animation baselines based on autoregressive video generation. (1) MIDAS~\cite{chen2025midas} trains a unified multimodal autoregressive model~\cite{bai2025qwen2} to fuse textual, audio, and visual information, thereby generating audio-aligned portrait animations. (2) TalkingMachines~\cite{low2025talkingmachines} builds upon CausVid~\cite{yin2025slow} distilling a bidirectional image-to-video (I2V) model into a few-step causal video diffusion model. Since neither method has been open-sourced, we conduct a visual comparison based on the demo videos provided on their respective project websites. Qualitative results are shown in Fig.~\ref{fig:exp_comp}, the top row shows visualizations of the baseline methods, while our corresponding results are presented in the bottom row. Visible artifacts in MIDAS stem from its limited visual detail modeling, as it decomposes frames into discrete tokens mixed with other modalities, compromising both texture fidelity and temporal coherence. TalkingMachines demonstrates strong visual stability and ID consistency, benefiting from adopting Wan2.1-14B as its base model, which provides a more powerful prior. In comparison, Knot Forcing achieves comparable performance with significantly lower computational cost.

\subsection{Compare with Causal Video Diffusion Models} 

We also compare with state-of-the-art causal video diffusion models~\cite{liu2025rolling, yang2025longlive, huang2025self, yin2025slow} for portrait generation. All of these models are few-step causal generators distilled from bidirectional teacher models.

\begin{figure}[t]
  \centering
   \includegraphics[width=0.95\linewidth]{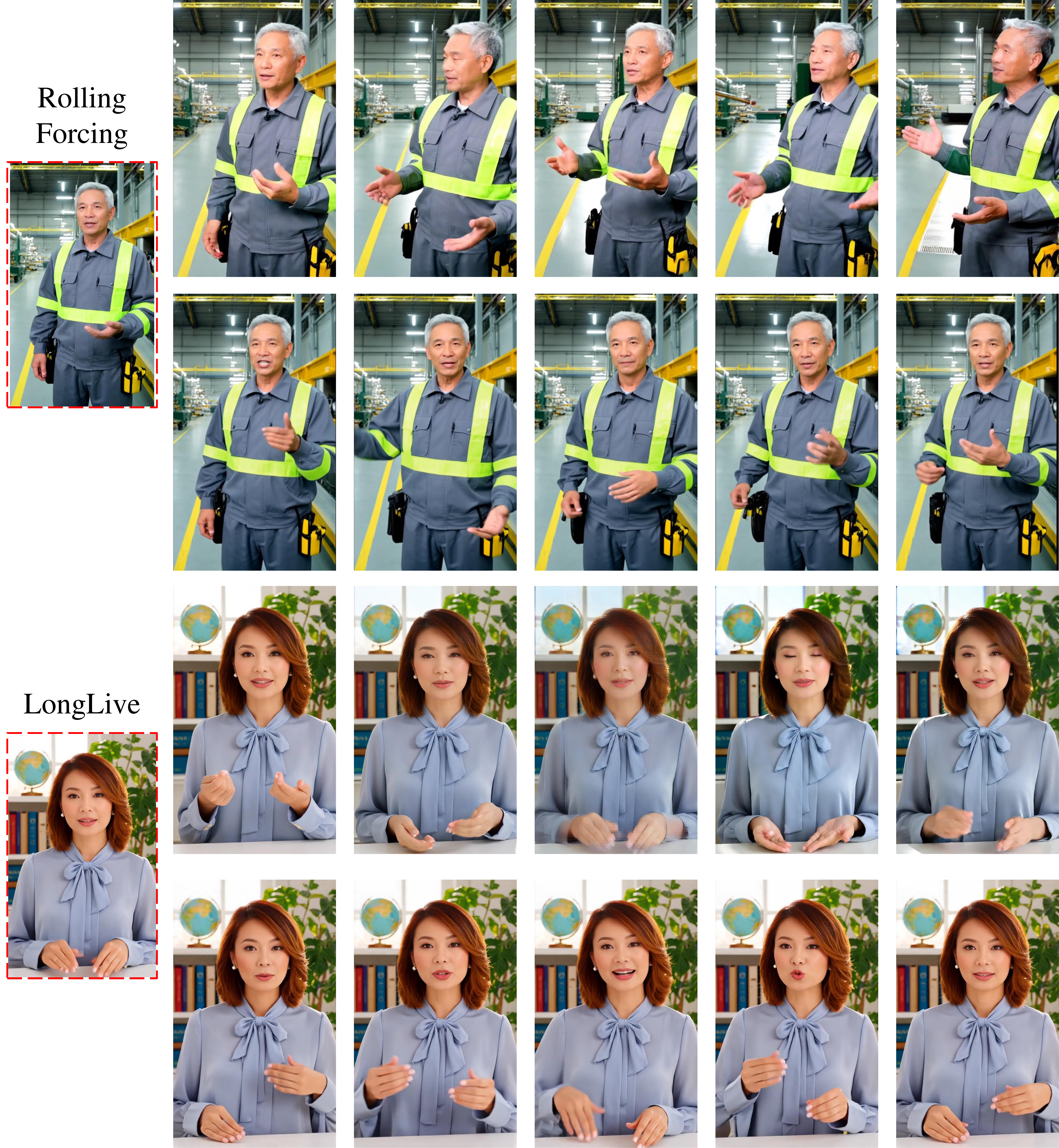}

   \caption{Comparison with causal-based video diffusion models. Top row shows results from the baseline methods, bottom row presents competing results generated by our method. The first frame from the videos generated by each baseline is used as the reference image for our model and is displayed in the leftmost column. Zoom in for better visual detail.}
   \label{fig:comp_caus}
   \vspace{-3mm}
\end{figure}

\noindent\textbf{Qualitative Comparison}. As illustrated in Fig.~\ref{fig:comp_caus}, we conduct visual comparisons with Rolling Forcing and LongLive on portrait animation to evaluate the quality, temporal consistency, and ID preservation of the generated videos. Since both methods are text-to-video models and cannot generate videos based on a specified reference image, we first use them to generate long videos from text prompts. We then leverage the first frame of generated video as the reference input for our model to produce comparable results. As can be observed, although both methods adopt attention sink to mitigate error accumulation, they still suffer from color drifting, identity shifts, and local distortions in long-horizon generation. In contrast, our method produces more stable results without error accumulation, preserving structural integrity without liquefaction. This is because (1) the introduced temporal knot establishes strong inter-frame temporal coherence and enables better imitation of the bidirectional teacher’s generation distribution, (2) while the global context running ahead provides a stable semantic target for future frames, preventing overall visual drifting.

\noindent\textbf{Quantitative Comparison}. We also conduct numerical comparisons with open-source autoregressive methods. We use VBench~\cite{huang2024vbench} quality metrics to evaluate generation quality on a set of 300 portrait-related prompts selected from MovieGen~\cite{polyak2024movie}. All videos are generated at $832\times 480$ resolution. For our method, we first generate a video using the bidirectional base model conditioned on the text prompt, then use its first frame as input for portrait animation in the evaluation. Quantitative results can be found in Tab.~\ref{tab:comparison}. As analyzed in Sec.~\ref{sec: Knot Forcing}, existing AR diffusion methods inevitably suffer from attention context mismatch, leading to uncontrollable visual distortions between consecutive frames, which severely degrades video stability and generation quality. In contrast, our method bridges the semantic gap across chunks, producing smoother, more stable, and higher-quality videos.


\subsection{Ablation Studies}

\begin{figure}[b]
  \centering
   \includegraphics[width=0.95\linewidth]{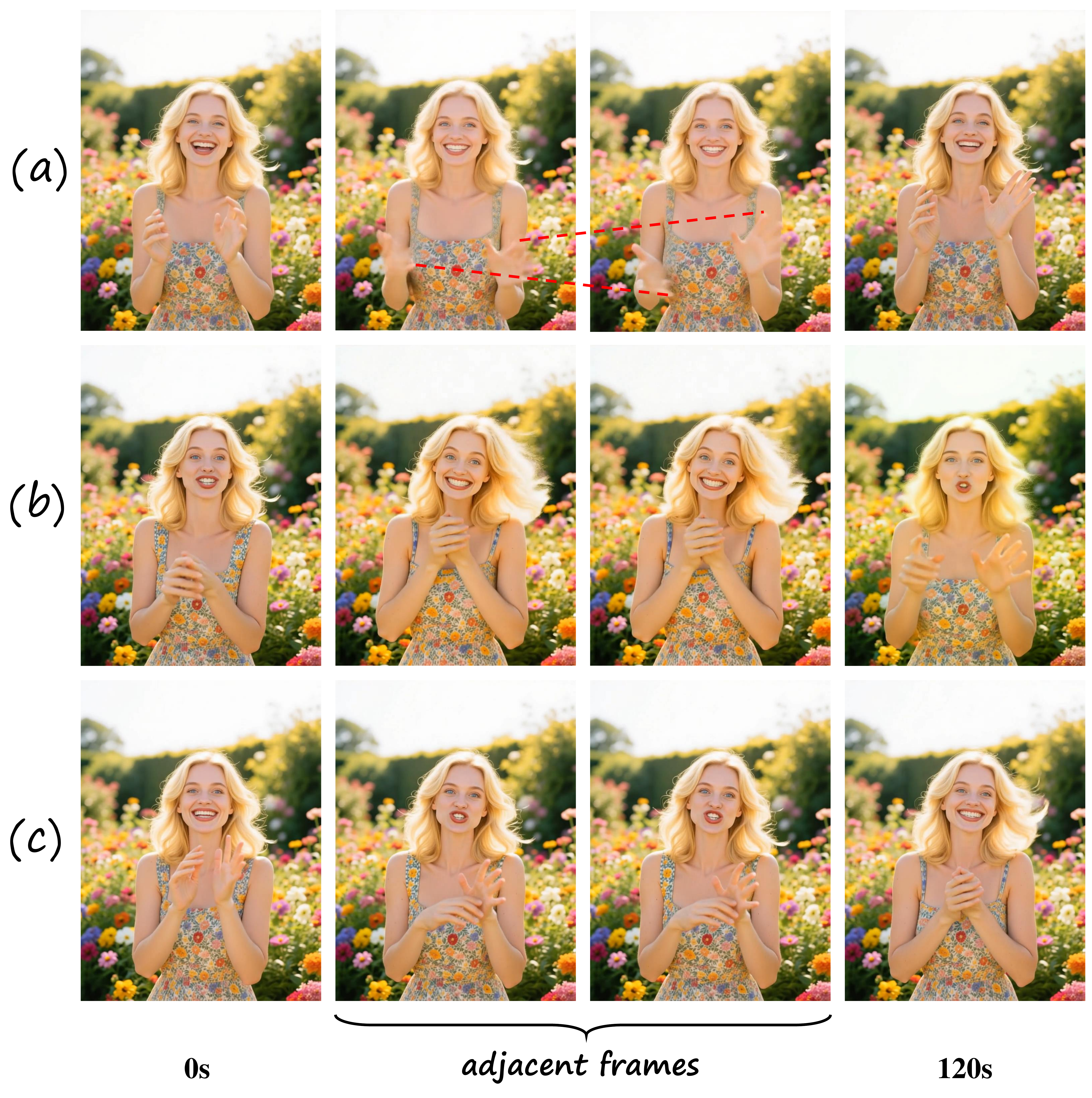}

   \caption{Visual effects of different components. From (a) to (c), three modules are integrated progressively: sliding window with global context,  temporal knot, and global context running ahead.}
   \label{fig:ablation}
   
\end{figure}

We conduct an ablation study to better understand the contribution of each designed component, and present the visual results in Fig.~\ref{fig:ablation}. From (a) to (c), we gradually add the three designed modules: (1) sliding window with global context, (2) temporal knot, (3) global context running ahead. As can be seen, when using only a sliding attention window with the reference image as global context ($1^{\text{st}}$ row), model tends to learn suboptimal solutions due to the relatively homogeneous data distribution in portrait animation. It simply replicates patterns from the reference image and focuses on intra-chunk semantic continuity. Consequently, when target motion pattern significantly deviates from the reference, the generation quality degrades noticeably, and undesirable motion jumps frequently occur. By introducing the temporal knot ($2^\text{nd}$ row), the semantic discontinuity between chunks is mitigated, and the model strengthens contextual coherence across frames. However, it tends to drift gradually from the global context over time, leading to noticeable semantic deviation in the overall video (last column). Further, the global context running ahead provides a stable semantic anchor for future frames (last row). The model maintains inter-frame continuity while staying on the correct semantic trajectory during rollout generation, effectively preventing visual drift.

\begin{table}[t]
  \small
  \setlength{\tabcolsep}{1.5pt} 
  \vspace{0.5em}
  \centering
  \resizebox{0.47\textwidth}{!}{
  \begin{tabular}{lc|ccccc}
      \toprule
      \multirow{3}{*}{Model} & \multirow{2}{*}{Throughput} & \multicolumn{5}{c}{Evaluation Scores $\uparrow$}   \\
    \cmidrule(lr){3-7}
      & (FPS) $\uparrow$ & \scriptsize Temporal   & \scriptsize Subject       & \scriptsize Background    & \scriptsize Aesthetic & \scriptsize Imaging  \\   &          & \scriptsize Flickering  & \scriptsize Consistency   & \scriptsize Consistency  & \scriptsize Quality   & \scriptsize Quality  \\
    \midrule
    CausVid~\citep{yin2025slow}    & 15.38 & 96.02 & 86.20 & 88.15 & 58.93  & 65.50 \\
    Self Forcing~\citep{huang2025self}  & 15.38 & 97.23 & 84.97 & 89.47 & 57.74  & 66.21  \\ 
    Rolling Forcing~\cite{liu2025rolling} & 15.79 & 96.91 & 90.89 & 93.01 & \textbf{63.11}  &	70.53 \\
    LongLive~\cite{yang2025longlive} & \textbf{20.70} & 97.82 & 91.80 & 93.42 & 62.56  & 72.01 \\
    Knot Forcing (Ours) & 17.50 & \textbf{98.50} & \textbf{94.05} & \textbf{96.26} & 63.09 &	\textbf{74.96} \\
    \bottomrule
  \end{tabular}
  }
  \caption{
    We compare Knot Forcing with autoregressive video generation models on generative quality. Best results are $\textbf{bold}$.
  }
  \label{tab:comparison}
  \vspace{-4mm}
\end{table}

\section{Conclusion}

This work proposes \textit{Knot Forcing}, a new causal video diffusion approach for infinite real-time portrait animation. To balance efficiency and identity consistency, we use a short sliding attention window and cache the reference image as a global context. To address two key issues in causal generation: temporal discontinuities between chunks and long-term visual drift, we introduce (1) temporal knot and (2) global context running ahead respectively. The former improves coherence by jointly denoising overlapping frames across chunks, while the latter dynamically updates the reference’s temporal position to guide long-term predictions. Together, they enable stable, high-fidelity, and temporally consistent animation. Future work will explore (1) theoretical analysis of the gap between causal students and bidirectional teachers, and (2) extending the framework to more general controllable generation tasks, such as world models and game environment simulations.
{
    \small
    \bibliographystyle{ieeenat_fullname}
    \bibliography{main}
}


\end{document}